# Can I trust my fake data- A comprehensive quality assessment framework for synthetic tabular data in healthcare

**Authors:**

Vallevik, Vibeke Binz[1, 2], Babic, Aleksandar[2], Marshall, Serena Elizabeth[2], Elvatun, Severin[3], Brøgger, Helga M. B. [2, 4], Sharmini Alagaratnam[2], Edwin, Bjørn[1,4], Veeraragavan, Narasimha Raghavan[3], Befring, Anne Kjersti[1], Nygård, Jan F.[3, 6].

**Author affiliations:**

[1] University of Oslo, Norway

[2] DNV AS, Norway.

[3] Cancer Registry of Norway, Norway

[4] Oslo University Hospital, Norway

[6] UiT The Arctic University of Norway, Norway

# Abstract

## Background

Ensuring safe adoption of AI tools in healthcare hinges on access to sufficient data for training, testing and validation. In response to privacy concerns and regulatory requirements, using synthetic data has been suggested. Synthetic data is created by training a generator on real data to produce a dataset with similar statistical properties. Competing metrics with differing taxonomies for quality evaluation have been suggested, resulting in a complex landscape. Optimising quality entails balancing considerations that make the data fit for use, yet relevant dimensions are left out of existing frameworks.

## Method

We performed a comprehensive literature review on the use of quality evaluation metrics on SD within the scope of tabular healthcare data and SD made using deep generative methods.

Based on this and the collective team experiences, we developed a conceptual framework for quality assurance. The applicability was benchmarked against a practical case from the Dutch National Cancer Registry.

## Conclusion

We present a conceptual framework for quality assurance of SD for AI applications in healthcare that aligns diverging taxonomies, expands on common quality dimensions to include the dimensions of Fairness and Carbon footprint, and proposes stages necessary to support real-life applications. Building trust in synthetic data by increasing transparency and reducing the safety risk will accelerate the development and uptake of trustworthy AI tools for the benefit of patients.

## Discussion

Despite the growing emphasis on algorithmic fairness and carbon footprint, these metrics were scarce in the literature review. The overwhelming focus was on statistical similarity using distance metrics while sequential logic detection was scarce. A consensus-backed framework that includes all relevant quality dimensions can provide assurance for safe and responsible real-life applications of SD. The right choice of metric is highly context dependent.

# Abbreviations

ACF - Auto Correlation Function

AE - Autoencoders

AI – Artificial Intelligence

AIA/AIR - Attribute inference attack /Attribute inference risk

ANOVA - Analysis of variance F-test

ARI - Adjusted Rand index

ATE - Average treatment effect (ATE)

AUPRC - Area Under the Precision-Recall Curve

AUROC/AUC - Area Under the Receiver Operating Characteristic Curve

BCE - Binary cross-entropy

CRN – Cancer Registry of Norway

DLA - Data labelling analysis

DM - Diffusion Models

DOP - distance to optimal point

DP - Differential Privacy

DS - Directional Symmetry

FDA - the U.S. Food and Drug Administration

FM - F-measure

FN – False Negative

FP – False Positive

FST - Fixation Index

FST - Fixation Index

GAN - Generative Adversarial Networks

GM - G-mean

IMDRF - The International Medical Device Regulators Forum

JS - Jaccard similarity

KLD - Kullback-Leibler Divergence

KS - Kolmogorov-Smirnov test

LIME - Local Interpretable Model-agnostic Explanations

LLM - Large Language Models

MAE - Mean Absolute Error (MAE) on predictions (cosine similarity calculated on MAE)

MAE - Mean absolute error for means and standard deviations of columns

MAEP - Mean Absolute Error Probability

MDR – Medical Device Regulation

MIA/MIR – Membership inference attack / Membership inference risk

ML – Machine Learning

MMD - Maximum Mean Discrepancy

NMI - normalized mutual information

NNAA - Nearest neighbor adversarial accuracy

NPV - Negative Predictive Value

PCA – Principal Component Analysis

PCD - Pairwise Correlation Difference

PEHE - Precision in Estimation of Heterogeneous Effect (PEHE))

PPC - Pairwise Pearson Correlation

PPV – Positive Predictive Value

PR curves - Precision-Recall Curve

PRESS - Predicted residual sum of squares

PW - Parzen window likelihood

PAGE 5Q2 - Predictive capacity

RMSE - Root Mean Square Error

RN - Required sample number

ROC - receiver operating characteristic curve

SaMD - Software as a Medical Device

SD – Synthetic data

SHAP - SHapley Additive exPlanations

std – standard deviation

STS - Short Time-Series Distance (STS)

SVM – Support Vector Machine

TATR - Train on Augmented Test on Real

TD – Training Data

THTR - Train on Hybrid Test on Real

TN- True Negative

TP – True Positive

TRTR - Train on Real Test on Real

t-SNE - t-distributed Stochastic Neighbor Embedding

TSTR - Train on Synthetic Test on Real

TSTS - Train on Synthetic Test on Synthetic

VIP – Variables importance in Prediction

PAGE 5

# 1 Introduction

Gartner predicts that by 2024, 60 % of the data used for development of AI and analytics projects will be synthetically generated[1]. Access to sufficient amounts of data poses a significant challenge for developing Machine Learning/Artificial Intelligence (AI) solutions in healthcare both due to the sensitive nature of healthcare data with resulting privacy and other regulatory issues[2, 3]. The datasets need to be big enough for effective training of AI tools[2]. Synthetic data (SD) derived from a purpose-built algorithm (generator)[4] that has been trained on a real dataset can facilitate both dataset augmentation and secure sharing of datasets[2, 5]. SD is currently used to broaden the understanding of healthcare data[6], for training and testing of AI tools and for experimentation in molecular research[7].

With the upcoming AI Act in the European Union (EU), the responsibility for systematic risk management will lie with designers, developers, deployers, manufacturers and distributors of AI systems[8-10]. There are obligations for safe, transparent, traceable, non-discriminatory and environmentally friendly use of AI tools in accordance with standards and legislation. The aim is to ensure proper functioning of AI tools during their lifecycle. Understanding the risk of an AI tool entails understanding the data this tool was trained and validated on, and how the tool should be used.

The popularity of deep learning methods like Generative Adversarial Networks (GAN), Autoencoders (AE) and more recently Large Language Models (LLM) and Diffusion models (DM) is growing as they appear to produce high quality SD[11]. However, use of SD gives rise to other concerns. The SD should clearly mimic the properties of the original data, but not become so similar that it poses a privacy risk [12] . Biases in the original dataset should not be amplified [5]. High computational complexity of these methods leads to substantial resource requirements[13].

**As such, using SD in high-risk AI applications related to human health warrants thorough data quality assurance.** There are many proposed, but no commonly recognized evaluation framework for SD[14]. Available evaluation metrics uses conflicting semantics, where authors allocate divergent meanings to the same word.

This paper introduces a conceptual framework for quality assurance of SD for AI applications in healthcare including a stepwise approach and a hierarchical taxonomy to clarify definitions and give an overview of important quality aspects. Our approach expands on existing frameworks to include the dimensions of Fairness and Carbon footprint, and steps that are necessary when preparing for clinical implementation.

Our framework is based on a comprehensive literature review on quality evaluation metrics. As medical data is often tabular, the scope was deep generative data synthesis for tabular data in healthcare. Mapping of quality metrics documented in the literature provides an overview of quality concerns and of the evaluation metrics most used to investigate them, and our framework provides a structure to understand which concerns the different types of metrics address.

Our ambition is to support a broad audience of potential users of SD in healthcare and provide a basis for further practical clinical implementation, with a common vocabulary to support cross-discipline communication and build trust in SD.

## 2 Material and methods

### 2.1 Process mapping and benchmark

Drawing on their experiences from SD at the Cancer Registry of Norway (CRN) and implementation of new technology at Oslo University Hospital, the team mapped the SD process from problem definition to clinical implementation of a trained tool. The quality assurance needs were identified at each stage, verified with available literature, and benchmarked with the SD process for a breast-cancer cohort at the Dutch National Cancer Registry (DNCR) through post-hoc interviews. The DNCR case was chosen for its similarity to CRN needs and because of a lack of available cases where SD was prepared for downstream clinical application.

### 2.2 Literature review – identification of relevant sources

This systematic review was conducted in accordance with the guidelines for the 'Preferred Reporting Items for Systematic Reviews and Meta-Analyses' extension for diagnostic accuracy studies statement (PRISMA-DTA)[15].

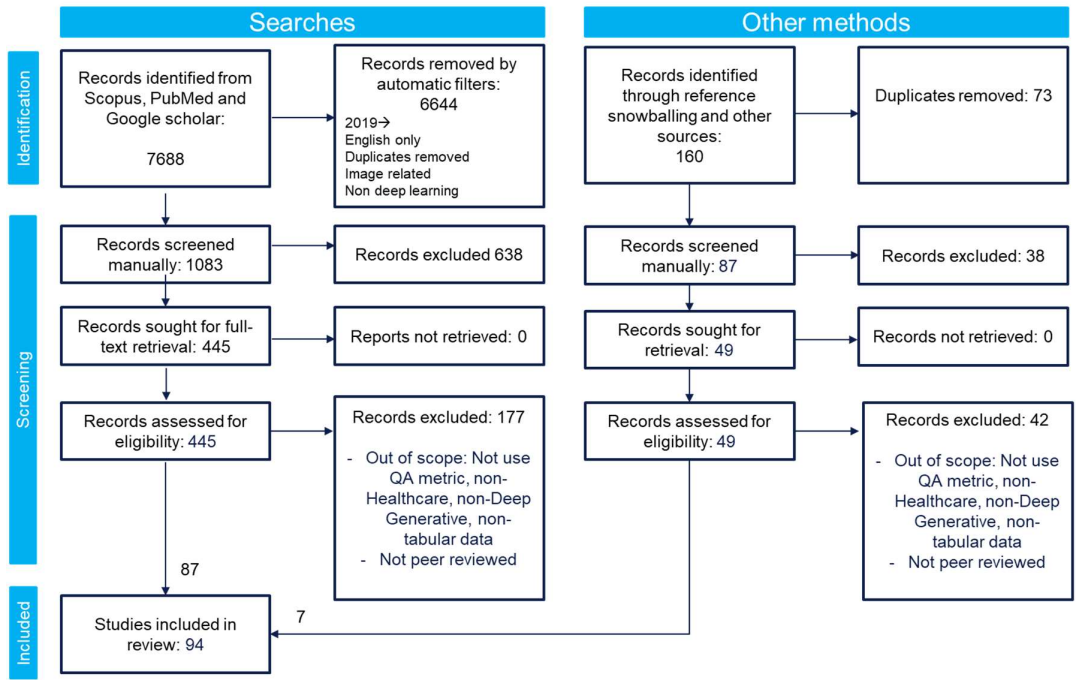

*Figure 1 PRISMA records*

Articles that report on quality evaluation or assurance perspectives in using SD for AI development in healthcare were sought.

Electronic bibliographic searches were conducted in Pubmed, Scopus and Google Scholar up to October 10[th] 2023. The search strings differed in the three engines due to syntax variations, see Table 1. An automatic filter to remove visual or image-based articles was used both for PubMed and Scopus. In addition, an automatic filter was applied on Scopus to single out the relevant technology.

*Table 1 Search strings for literature review.*

| Engine | Search location | Search terms |
|---|---|---|
| PubMed | Title and Abstract | ("synthetic data*" OR "synthesized data" OR "synthetically generated " OR "synthetic patient data" OR "artificial* data*" OR "augmented data*") <br><br> AND <br><br> (assess* OR assur* OR perform* OR quality* OR validat* OR verif* OR evaluat* OR framework*)) <br><br> AND <br><br> (health* OR patient* OR lifescience OR life-science OR "life science" OR medic* OR biomedic* OR diagnos*) <br><br> NOT <br><br> ( imag* OR video OR vision OR audio OR quantum OR radar OR mri ) [Title/Abstract] |
| Scopus | Title | (synthetic AND data* OR gan OR autoencod* OR "diffusion model*") <br><br> AND |
|  | Title, abstract and keywords | (performance OR validat* OR quality OR assurance OR evaluat* OR framework OR review OR survey) <br><br> AND <br><br> (health* OR patient* OR medicine OR lifescience OR life-science OR "life science")) <br> AND NOT <br><br> ( imag* OR video OR vision OR audio OR quantum OR radar OR mri ) <br><br> AND NOT <br><br> ( "Generative* model" OR gan OR autoencod* OR variational* OR "bayesian neural network" OR "autoregressive model" OR "bolzmann machine" OR "deep belief network" OR "diffusion model" )) |
| Google scholar | Title | synthetically generated data (healthcare OR health OR patient) OR assurance OR performance OR quality OR validation OR evaluation |
| Other |  | Additional articles were identified through manual searches of bibliographies and citations until no further relevant articles were identified. |

Two investigators independently screened titles and abstracts and selected relevant articles for full-text review. Extra searches were performed for clinical validation and monitoring articles.

## 2.3   Inclusion criteria

The time range was set from 2019. Articles that evaluated quality of tabular healthcare data that had been generated by deep generative methods were included, based on the classification of deep generative methods from Hernandez, Epelde [16].

Only reviews that applied quality metrics were included. The articles that proposed evaluation frameworks without applying metrics were kept separate to be used for an analysis of existing frameworks (chapter 3.1).

Tabular healthcare data consist of columns of variables of mixed types, excluding continuous time-series data from sensor measurements with only one variable.

In the initial screening, twelve frameworks for quality evaluation were identified. Some were not included in the literature review but were used for mapping existing frameworks to build on. Disagreement regarding study inclusion was resolved by discussion with a third investigator.

### 2.4    Exclusion criteria

Articles were excluded if they were not peer reviewed, not written in English or were a short version of another retrieved publication. Preprints that had been accepted for conference proceedings were included. Articles from predatory journals were excluded [17], as were PhD-theses.

### 2.5    Data extraction and quality assurance

One investigator evaluated and retrieved information from the article about quality assurance metrics that were documented. These metrics were cross mapped with the stages in our designed process and quality metrics in our framework. In the cases where one article encompassed several data modalities, the investigators sought to include only the tabular data metrics where separation was feasible.

A separate investigator reviewed the mappings. By uncertainties or disagreements, three or more investigators discussed as a group to reach consensus.

### 2.6    Data synthesis and analysis

The metrics were divided into subgroups of quality dimensions according to what authors claim the metric should measure. Some interpretation was necessary when diverging semantics were used. The groupings were quality assured by two separate investigators, diverging conclusions were discussed for consensus in a wider group of three or more investigators.

### 2.7    Ethics

No new person data was gathered from humans.

## 3    Results

### 3.1    Diverging taxonomy and evaluation criteria

Our initial literature screening identified twelve frameworks for SD quality evaluation (details in chapter 2) with diverging taxonomy. Notably, the term utility is widely used to denote overall fitness for use, encompassing both similarity and downstream outcome predictions. The term is often used colloquially in the context of the  "utility-privacy trade-off" [18]. However, authors like Hernandez, Epelde [16]  make a distinction between statistical similarity (resemblance) and conclusions drawn from the data (utility). Dankar, Ibrahim [19] use both fidelity and utility intermittently to denote overall fitness for use but distinguish between statistical similarity (broad utility) and application specific performance (narrow utility). Alaa, Van Breugel [20] on the other hand, use fidelity specifically for evaluating the similarity of distribution coverage.

Some authors introduce novel quality measures by combining existing metrics [14, 21, 22]. This simplifies comparison but may challenge application-specific adaptation and reduce transparency in the trade-off between quality dimensions.

## 3.2 A proposed taxonomy for quality dimensions in the evaluation of SD

In alignment with the EU definition of data quality, our proposed framework builds on a multi-dimensional and context specific perception of quality: "fitness for purpose for users' needs (…) and that data reflect the reality, which they aim to represent"[23]. This interpretation is in line with the philosophical direction pointed out by Pirsig [24], where quality is the dynamic equilibrium between different aspects of existence.

Our taxonomy comprises five dimensions for evaluating synthetic data: Similarity, Usability, Privacy, Fairness, and Carbon footprint and computational complexity (hereafter Carbon footprint).

Figure 2 includes our definitions and other terms used for the concepts and illustrates the hierarchical structure. We aim to follow the Mutually Exclusive and Collectively Exhaustive (MECE) principle.

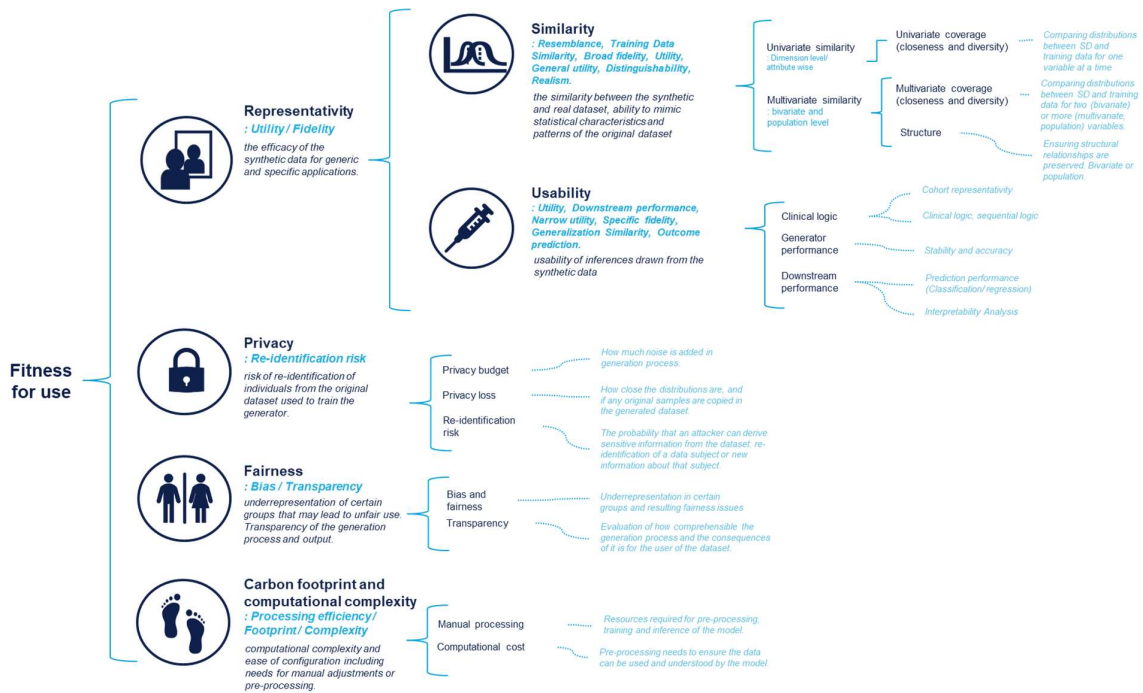

*Figure 2: Proposed taxonomy of quality dimensions with definitions and other terms used for the same concept.*

To mitigate confusion arising from varied use of terms like Utility and Fidelity, we propose adopting the term Representativity as a superset of Similarity and Usability. Representativity is evaluated through SD comparison with the training dataset (Similarity), or by comparing the results generated by the SD with results generated by the training dataset (Usability). Similarity metrics address statistical similarity in terms of coverage – closeness or diversity - of data distributions, as well as the structure and patterns in the data. Univariate comparisons are employed to compare a specific variable in the SD with the same variable in the training data, while multivariate evaluations encompass statistical properties of more than two variables. Usability or the "usefulness to solve a given problem"[25], assesses whether the data is realistic enough that downstream results are equally valid in the real world [26]. The training data should be representative for the intended cohort and to perform well in a specific application, case relevant clinical logic should be preserved in the SD.

Even from fully SD there may be a risk of deriving personal sensitive information [27]. The generator could overfit and copy original data samples, or the generated data can be statistically close enough to the real data to allow inferences to be made. Techniques like Differential Privacy (DP) may be employed during the generation process, but balancing privacy and other dimensions like similarity or usability requires careful consideration [28].

Bias in training datasets may lead to algorithms unfairly penalizing minority groups within a population [29]. Bias amplification is a concern in SD [30] if the generation method fails to capture the distribution of the real data subgroups effectively across all protected attributes[31]. Fairness concerns emerge when decision making is informed by biased datasets, and renders a need for transparency and comprehensibility also to non-data science experts[29].

The substantial energy consumption of compute-intensive generative AI models give rise to significant environmental concerns[32]. Carbon footprint and computational complexity encompasses resources required for preprocessing, training and inference from a model to produce the SD. Resource needs are measured in model size, computation time, type of energy sources and storage methods employed, but also manual processing needed for complex models. These are measures of the cost but also the sustainability of a method and its resulting dataset.

Table 2 summarises diverging terminology used in evaluation frameworks, emphasizing the need for a standardized approach.

*Table 2 Examples of evaluation frameworks and their quality dimensions. No current framework covers all five quality dimensions.*

| Year | Proposed evaluation frameworks | Similarity | Usability | Privacy | Fairness | Carbon footprint |
|---|---|---|---|---|---|---|
| 2023 | Hernadez, Epelde [33] | «Resemblance» | «Utility» | «Privacy» | - | *Mentioned, not included |
| 2022 | Yan, Yan [34] | "Data utility: resemblance" | "Data utility: outcome prediction" | «Privacy» | - | - |
| 2022 | Dankar, Ibrahim [19] | "Broad Utility measures: Attribute, bivariate and population fidelity" | "Narrow Utility measures: application fidelity" | - | - | - |
| 2022 | Figueira and Vaz [11] | "Compare statistics" | "Machine learning efficacy" | | | |
| 2022 | Alaa, Van Breugel [20] | "Utility": "Fidelity" & "Diversity" | - | «Authenticity» | - | - |
| 2021 | Hernandez, Epelde [14] | «Resemblance» | «Utility» | «Privacy» | - | *Mentioned, not included |
| 2021 | Dankar and Ibrahim [35] | "Global utility measures" | "Analysis-specific measures" | - | - | - |
| 2020 | El Emam [36] | «Utility» | «Utility» | *Mentioned, not included | "Bias and stability assessment" | - |
| 2020 (preprint) | Arnold and Neunhoeffer [37] | "Training data similarity" | "Generalisation Similarity" | - | - | - |
| 2020 (preprint) | Djolonga, Lučić [38] | "Discrepancy" | - | - | - | - |
| 2019 | Alqahtani, Kavakli-Thorne [39] | * Provides a list of metrics, no split between Similarity or usability but both dimensions are covered. | | | | |
| 2018 | McLachlan, Dube [40] | "Realism" | - | - | - | - |

## 3.3 Gaps in the use of quality dimensions in the literature

A literature review on deep generative methods for synthetic data evaluation in tabular healthcare data from 2019-2023 identified 94 articles, yielding 616 metrics that were categorized according to the quality dimension and process stage. Details are provided in section 2 Material and methods and Appendix B Additional results from the literature review.

Representativity metrics (similarity and usability) dominated comprising over 80 %, while privacy, fairness and carbon footprint metrics collectively accounted for less than 20 % of the recorded metrics (Figure 3a). Only one article documented metrics in all five dimensions, 88% of the articles only used one, two or three quality dimensions. (Figure 3b). Although fairness and transparency are growing topics in AI and SD (Figure 3 c), only six articles were found that applied these. Four articles investigated bias in the training dataset (5/616 metrics), and two articles evaluated fairness in a downstream application (5/616 metrics). Carbon footprint and computational complexity was registered in 28 of the 94 articles (32/616 metrics), yet these concerned pre-processing resource needs (22/616) or computational resource needs like size, training time or storage (10/616) rather than the actual carbon footprint.

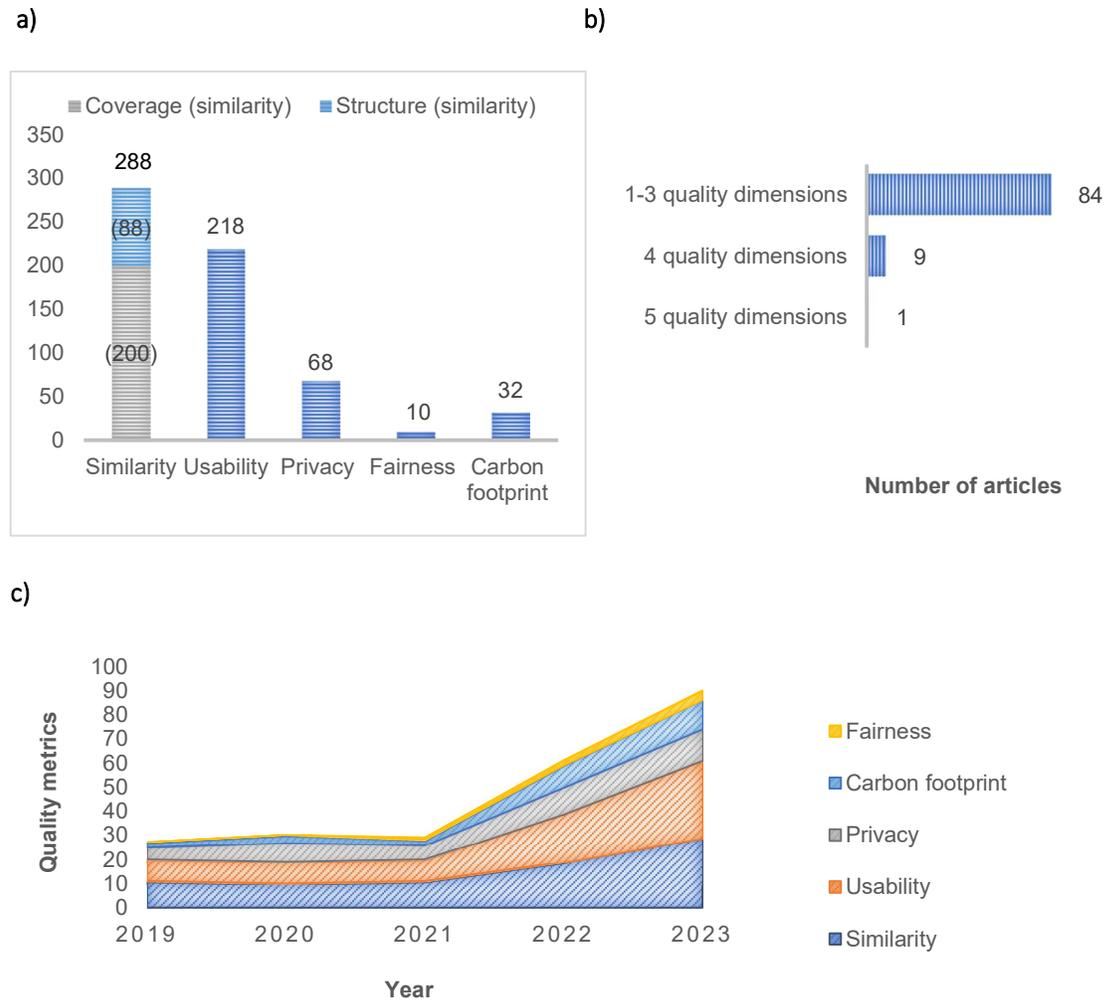

*Figure 3 Quality dimensions results: a) Number of articles that document use of the different quality categories in the literature review, b) Percentage of articles covering one- three, four or five quality dimensions and c) Types of quality metrics pr year.*

288 similarity metrics in total were found in 78 of the 94 articles, amounting to an average of 3,7 similarity metrics per article. Some applied more than 10 similarity metrics within one article [14, 41]. 200 metrics compared similarity of distribution coverage (distance or diversity) and 88 compared similarity in structural relationships.

71 of the 94 articles tested data usability by benchmarking downstream performance on models trained on the data. 3 of these tested policy recommendations from reinforcement learning agents, 7 used regression models for predictions and 61 used classifiers for predictions. 158 registered classification model performance metrics amounted to 23 unique metrics. Most were calculations based on confusion matrix values. Reporting up to 8 of these metrics in one article will not necessarily add value.

Two of the six articles that evaluated privacy budget through Differential Privacy (DP) also applied post hoc evaluation metrics to measure privacy loss or privacy risk [42, 43].

### 3.4 A stepwise approach to quality assurance

Our proposed framework provides a stepwise approach to assess synthetic data quality, aligning with EU medical device regulation (MDR)[44] and FDA guidance for AI/ML applications[45]. Domain knowledge, a crucial element, can be automated where possible to ensure both quantitative and qualitative evaluations. A complete evaluation will typically contain both quantitative and qualitative elements. Certain quantitative metrics can be built into the pipeline and discard samples below a defined quality threshold, as suggested by Alaa, Van Breugel [20]. Domain knowledge is crucial and some aspects may be automated, like in logical rules engines[46].

Figure 4 outlines the process, distinguishing between upstream data production and downstream data application.

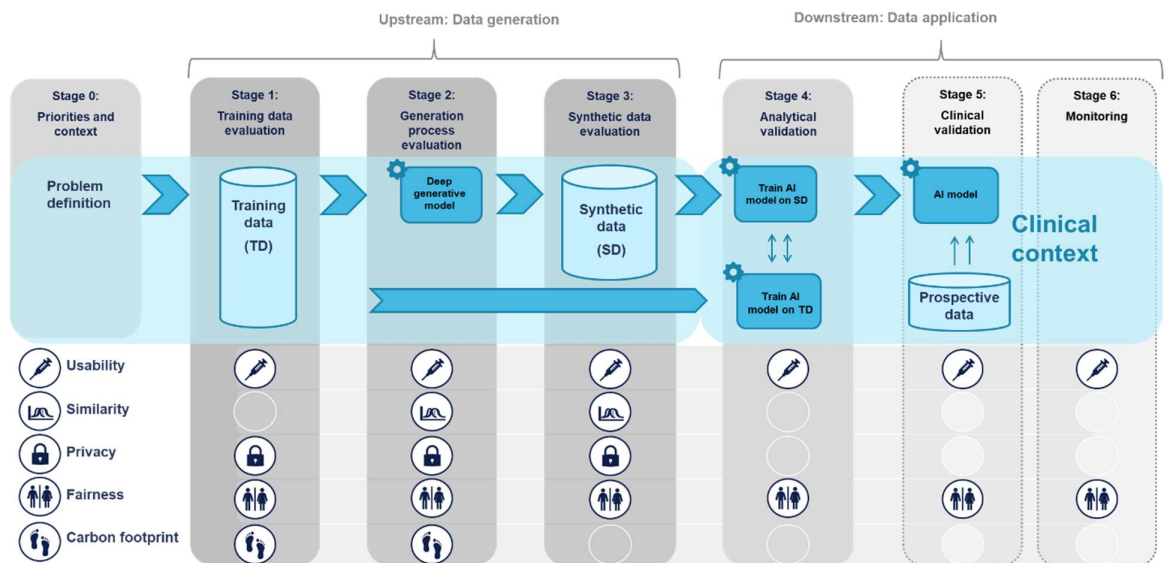

*Figure 4 The process of quality assurance follows the entire pipeline of SD generation and use.*

Table 3 summarizes quality considerations to be made at each stage in the process, and the methods identified in the literature. The stages are briefly commented in the following.

#### 3.4.1 Stage 0: Priorities and context

The context for use of the SD and the issue it should solve is central to tailoring the quality assurance and deciding the relative importance of the different quality dimensions.

PAGE 15

### 3.4.2 Stage 1: Training data quality evaluation

The validity of SD inferences depends on the quality of the training data. "Record completeness"[5] concerns whether the dataset is technically suitable for use and results in pre-processing to counter missing data, outliers etc. "Case completeness"[5] is necessary for valid inferences downstream. The training data population must be representative and relevant, preferably diverse enough to yield robust results. Specific steps in the clinical pathways or data capture process may influence the relevance of the data for other uses[47], i.e. a measurement value must be interpreted in the correct context depending on whether it was obtained before or after an intervention. Selection bias should be investigated to identify and handle under-representation of minorities [2].

### 3.4.3 Stage 2: Generation performance evaluation

Using literature to guide generator selection may result in conflicting conclusions[19] as studies use different datasets and metrics to evaluate their methods' performance[16]. There is no consensus on objectively classifying quality of generators[48]. Common challenges in SD generation encompass algorithm convergence, mode collapse, mode invention, density shifts[20] or noisy data, copying of training data, inadequate coverage[49], and bias amplification[30]. SD generators may be optimized to fit certain data modalities like images, text, or structured data. Tabular data are particularly challenging due to mixed feature types, imbalanced proportion of outcomes and temporal dependencies[46]. Complex generators may become computationally unsustainable.

### 3.4.4 Stage 3: Synthetic data evaluation

Representativity can be evaluated by statistical comparison with real datasets or medical knowledge (such as disease prevalence), expert evaluations and sample-level logical rule engines[46]. Rigorous testing of variables crucial for the intended application is advised.

### 3.4.5 Stage 4: Analytical validation

Stages 4, 5 and 6 assesses performance and inferences drawn from models for specific downstream applications that are trained on SD. Stage 4 is in the analytical setting often using split-sample validation [45].

The prevalent method benchmarks performance between a model Trained on Synthetic and Tested on Real data (TSTR) with the baseline of a model Trained on Real and Tested on Real (TRTR), alternatively using hybrid or augmented datasets (TATR/THTR). Variations exist like Train on Real Test on Synthetic (TRTS), Train on Synthetic Test on Synthetic (TSTS) etc. A larger dataset may yield better results due to size rather than the quality of medical inferences, so caution is warranted when benchmark dataset sizes are significantly different.

### 3.4.6 Stage 5 and 6: Clinical validation and Monitoring

The goal of clinical validation in MDR [44] is to demonstrate to regulators that the device will work as intended, for the target population and in the clinical context (Stage 5), stepping out of the laboratory setting of Stage 4 and into a real-world setting [45].

Following implementation, performance monitoring (Stage 6) feeds into Post Market Surveillance.

*Table 3 Quality considerations in each stage and methods for quality assurance used in the literature. Further details in Appendix B Additional results from the literature review.*

| | Considerations | Methods and metric examples | Freq |
|---|---|---|---|
| | **Stage 0** | | |
| | What is the intended use of the dataset and the needs in terms of sharing, recreating the original data or generating new outlier scenarios? How should the quality dimensions be weighted accordingly?<br>Are any variables in the dataset expected to be more important than others for the expected use? | | NA |
| | **Stage 1** | | |
| 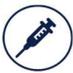 | Clinical logic:<br>Does the original dataset represent well the patient group it is intended for? Does the dataset provide a good representation of reality? (Case-completeness) | Qualitative expert evaluation. Decide on relevant variables. [41, 50] Compare with population statistic or medical knowledge. | 2/94 articles |
| 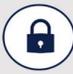 | Is the inherent privacy risk in the original data too high for direct processing? The need for de-identification measures should be carefully considered and balanced with utility.<br>DNCR | Qualitative expert evaluation. [51] | 1/94 articles |
| 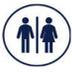 | Bias or imbalance: Were any groups underrepresented in the dataset before or after pre-processing. | IR - imbalanced ratio, Balancing, downsampling, upsampling, data imputation. [50, 52-54] | 4/94 articles |
| 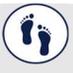 | Resources needed for preprocessing (record level data quality)<br>DNCR | Evaluate pre-processing needs in generators:<br>Does it require labels (Annotations), Can it handle mixed data types, does it capture correlated and temporal Information?<br>Data curation - Outlier detection (z score), remove outliers, data characterization, detection of problematic and inconsistent fields, de-duplication. Missing data imputation/ remove patients with missing features. Normalization and padding. Reduce number of node labels for the learning algorithm. [5, 33, 41, 51, 52, 55-70] | 23/94 articles |
| | **Stage 2** | | |
| 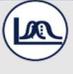 | Monitor similarity of the latent structure with the training data.<br>DNCR | Latent cluster analysis, Latent space representation (LSR), Log-cluster metric $U$, Category coverage (CAT), Log-Cluster Metric (LCM). [34, 71-76] | 7/94 articles |
| 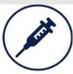 | Is the generator fit for use? Monitor generation performance.<br>DNCR | Ablation study, Convergence loss, Correlation loss, Discriminator testing, critic, generator and classifier loss - Visual plots, Stability of the training regime,<br>Predict the presence of potential medical codes.<br>[25, 41, 51, 58, 70, 73, 77-82] | 12/94 articles |

| | Considerations | Methods and metric examples | Freq |
|---|---|---|---|
| 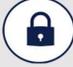 | How well does the generation method preserve privacy? Define privacy budget and evaluate trade-offs.<br>DNCR | Differential privacy guarantees - evaluate Epsilon values. (Privacy budget) [43, 70, 83-86] | 6/94 articles |
| 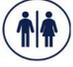 | How does the generator handle biases? | Investigate literature. | 0/94 articles |
| 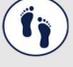 | How is the complexity of the generator in terms of pre-processing needs, tuning of hyperparameters and computational power needs?<br>DNCR | Carbon footprint: model footprint (size), memory usage, average runtime - training and inference time (generation time, execution time). [57, 58, 71, 74, 87-90] | 8/94 articles |
| | Stage 3 | | |
| 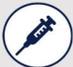 | Is the synthetic patient realistic?<br>DNCR | Expert evaluations, rule engines.<br>[34, 50, 76, 79, 83, 86, 91-93] | 9/94 articles |
| 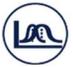 | Univariate level similarity: How similar are the variables (columns) in the SD compared to the same variables in the training data in terms of closeness and diversity?<br>DNCR | Basic statistics (Summary Statistic Comparison)[41, 61, 62, 64, 68, 82, 86, 93-99]<br><br>Dimension-wise probability (visual plots, frequency comparisons, Bernoulli success probability, Site Frequency Spectrum (SFS), isomap plots, marginal distributions, scatter plots, probability distribution) [14, 34, 41, 42, 50, 56, 58, 62, 64, 67, 68, 73, 76, 78, 79, 81-83, 90-92, 99-107]<br><br>Statistical tests (Distance or divergence tests - KS, Jensen Shannon, HD, Wasserstein, KLD, cosine distance- Jaccard similarity, Mann-Whitney U-test, Pearson's Chi-square ($\chi^2$), MMD, MAEP, NNAA, Precision and recall ,PW, ANOVA, RMSE, three sigma, student t etc.) [14, 20, 33, 43, 51, 53, 55, 58, 59, 62, 64, 67, 70, 72-75, 77, 79, 85, 96, 98-101, 103-105, 107-117]<br><br>Survival analysis (kaplan-meier divergence, Optimism, Short-sightedness)[77]<br><br>Visual plots (f.ex. t-SNE)[64] | 63/94 articles |
| | Multivariate level comparison of distribution coverage:<br>How similar are the distributions of two or more variables in the datasets in terms of coverage?<br>DNCR | Statistical tests (MMD, MMAE, Mean-log likelihood, Euclidian distance, peacock test (multi-KS), Wasserstein distance, Multivariate HD, Jensen Shannon, medical concept abundance, ANOVA, freq. of propensity scores) [5, 34, 42, 54, 57, 58, 62, 69, 75, 90, 98-100, 108, 110, 118, 119]<br><br>Diversity metrics (alpha and beta-diversity, std of gaussian noise, support coverage, k-means, RN)[56, 69, 74, 94, 113]<br><br>Longitudinal metrics (Shortest path kernel, Weisfeiler-Lehman subtree kernel)[69]<br><br>ML based methods (Data Labelling Analysis (DLA)/ Binary classification, Propensity MSE, Propensity score, NNAA[14, 33, 42, 57, 59, 64, 71, 75, 83, 87, 88, 99, 101, 106, 109, 112] | 33/94 articles |
| | Multivariate level comparison of dataset structure: How similar are the underlying patterns, clusters and correlations in the two datasets?<br>DNCR | Correlation (visual plots or coefficients (Pearson, Spearman). Heatmaps, matrices, correlation coefficients - Pairwise Pearson Correlation (PPC), Pairwise Correlation Difference (PCD) - co-variance matrices, Kendall's Rank correlation) [34, 55, 57, 58, 98, 101, 109] [14, 33, 42, 43, 53, 54, 56, 59, 61, 68, 69, 72-74, 76, 86, 91, 94, 96, 99, 100, 103-105, 107, 111, 115, 116] | 52/94 articles |

| Considerations | Methods and metric examples | Freq |
|---|---|---|
| | Dimensionality reduction (PCA, UMAP, ARI and NMI scores of clustering results, visual plots, t-SNE, K-means clustering, covariate plots) [14, 56, 57, 61, 63, 66, 87, 95, 108, 111, 112, 117, 120] | |
| | Feature importance (feature correlation, ranking) [84, 108, 121] | |
| | Longitudinal (Trend correlation, SD, Cosine similarity, median trajectories, mean vectors plot, STS, MAE of ACF, Disease progression patterns, KS on latent temporal statistics, HD of transition matrices, Directional symmetry) [42, 56, 59, 62, 86, 93, 98, 107, 111, 120] | |
| | ML based methods (Association rule mining, Dimension-wise prediction (DWP), Frequent Association Rules (FAR)) [70, 76, 90, 101, 102, 115] | |
| | Epidemiological (Kaplan–Meier (K-M) curve, Cox regression, Hazard ratios, Odds Ratio)[87, 117, 122, 123] | |
| Are the datapoints too similar to the original? (copies)  DNCR | Overfitting[122] | 22/94 articles |
| | Distance metrics (NNAA, RMSE, Euclidean distance, Hausdorff distance, Hamming distance, Cosine similarity)[14, 33, 34, 61, 73, 87, 88, 98, 106-108, 120, 122, 124] | |
| | Authenticity (Duplicate count, Mann-Whitney U-test, RSVR, KS)[20, 58, 72, 99, 104, 105] | |
| What is the risk of an attacker being successful in inferencing personal sensitive data from the SD? | Privacy Attack Simulations (Membership Inference Attack, Attribute Membership Attack, Meaningful identity disclosure risk, Re-identification attack)[14, 25, 27, 33, 34, 42, 43, 59, 62, 64, 73, 74, 76, 93, 100-102, 107, 117, 122, 124, 125] | 23/94 articles |
| Are any groups underrepresented in the dataset and are key outcome variables different for these groups? | Fairness metrics (Log Disparity, Time-Series Disparity) [31] | 1/94 articles |
| Stage 4 | | |
| Compare performance of the model trained on SD with the same model trained on another dataset, f. ex. training data.  DNCR | Classification performance (Accuracy/average accuracy/balanced accuracy/classification accuracy/prediction accuracy/mean accuracy, Precision – recall, PR curves/ Area under curve (ROC and AUC), Area Under the Precision-Recall Curve (AUPRC), Area Under the Receiver Operating Characteristic Curve (AUROC), Binary cross-entropy (BCE) loss, Brier score, Confusion matrix - TP, TN, FP, FN, DOP (distance to optimal point), Error rate, F1-score/ Macro F1-score / Weighted F-1 score (w-F1) / Abf F1/ F-measure (FM)/F-score, G-mean (GM), Granger causality, Hamming Loss, Jaccard Similarity score, Mean Absolute Error (MAE) on predictions, NPV, PPV, Sensitivity, Specificity, SVM agreement rate, SVM misclassification rate, The Brier score, Total cost of model misclassification, Wilsons method for confidence intervals, Youden's index)[34, 55, 57, 80, 98] [14, 25, 33, 42, 43, 50, 52-54, 56, 59, 60, 63-65, 67-72, 74, 77, 78, 81, 82, 84-89, 91-93, 95, 99, 102, 106, 108-110, 113, 114, 116, 119, 120, 123, 125-132] | 71/94 articles |
| | Regression performance (Goodness of fit (R2), Accuracy, AUROC + AUPRC, Kendall rank coefficient, Sqrt(Error_PEHE), Error in Average treatment effect (ATE) Predicted residual sum of squares (PRESS), Prediction error, Predictive capacity (Q2), R2 score, RMSE)[66, 75, 97, 104, 108, 118, 120] | |

| | Considerations | Methods and metric examples | Freq |
|---|---|---|---|
| | | Policy learning accuracy (Suggested drug combination, Relative Frequencies of Actions Taken (plots), top-k accuracy indicators) [73, 97, 107] | |
| | | Interpretability Analysis (SHAP and LIME (xAI) testing logical rules (LLM), Top-N recall and precision values, VIP analysis comparison, Feature correlation) [34, 66, 72, 83, 86, 90, 104, 109, 128, 132] | |
| | | Results Ranking (Concordance index (C-Index), Synthetic Ranking Agreement (SRA), "ground-truth" ranking of models) [20, 77, 84] | |
| 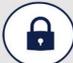 | Is there a privacy risk in inferencing sensitive information downstream? | Differential privacy in downstream model. [42] | 1/94 articles |
| 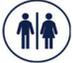 | Do minority populations achieve the same prediction performance? | Fairness evaluation (Demographic parity, Equalized odds, Overall accuracy equality.)[64] | 1/94 articles |
| | **Stage 5 & 6** | | |
| 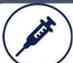 | What is the robustness of the model trained on SD? | Performance test on real data in a clinical context. | 0/94 articles |
| | Is the model performing over time? | Data or model drift. | |
| 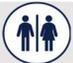 | Do minority populations achieve the same prediction performance? | Fairness metrics. | 0/94 articles |
| | Does the model perform for minority groups over time? | | |

## 3.5 DNCR benchmark

The framework was applied post hoc to the data generation process from a breast cancer cohort at the Dutch National Cancer Registry (DNCR). The assurance aspect they covered is illustrated with a ⬚DNCR in Table 3 and in Figure 5.

Stage 0: The intention was to share a generic dataset showcasing the data available to allow for initial exploration and software development without the need to see real patient records. Privacy was deemed the most important factor, followed by structural similarity.

Stage 1: The data was de-identified prior to generating the SD to reduce privacy risk by grouping variables and removing outliers.

Stage 2: When testing generators recommended in the literature, different privacy budgets (using differential privacy) were tested. Computational complexity was considered.

Stage 3: Privacy preservation was prioritized over realistic synthetic patients. Univariate similarities, pairwise correlations and privacy loss was tested.

Stage 4: Downstream performance was measured through ML prediction tasks and survival analysis. The results from these were very dependent on what variable they tested for.

Stage 5 & 6: Not relevant as there was no planned specific downstream use.

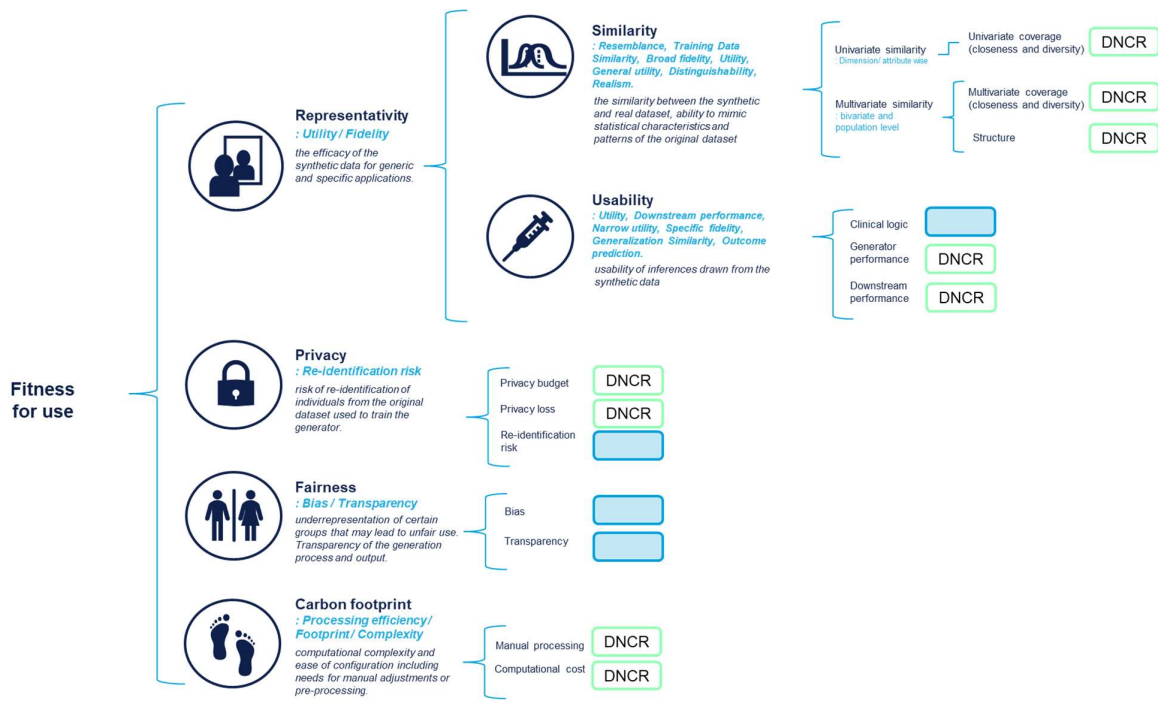

*Figure 5 The quality considerations covered in the DNCR case. The squares marked DNCR indicates the considerations that were included in the DNCR case, the empty squares were not included.*

Figure 5 illustrates the DNCR case covered a broad array of quality concerns but did not include clinical logic evaluation of the training data nor the SD, privacy simulation or any fairness perspectives. Considerations were done on computational cost and footprint, but no calculation of carbon footprint was made. Further details in Appendix A Benchmarking the framework with a case from the Dutch National Cancer Registry.

## 3.6   Gaps in the process stages reported in the literature

Half of the articles originated from technology focused journals, the other half from healthcare specific publications. 60 % had the main purpose of proposing a new generators or generation processes, 24 % sought to validate or test generators, while 16 % proposed or validated evaluation metrics.

A total of 616 metrics were mapped across different stages, revealing a predominant focus on Stage 3 data evaluation (82 of the 94 articles) and Stage 4 analytical validation (71 of the 94 articles). No articles documented clinical validation of an AI tool trained on SD (Stage 5), nor monitoring (Stage 6), see Figure 6.

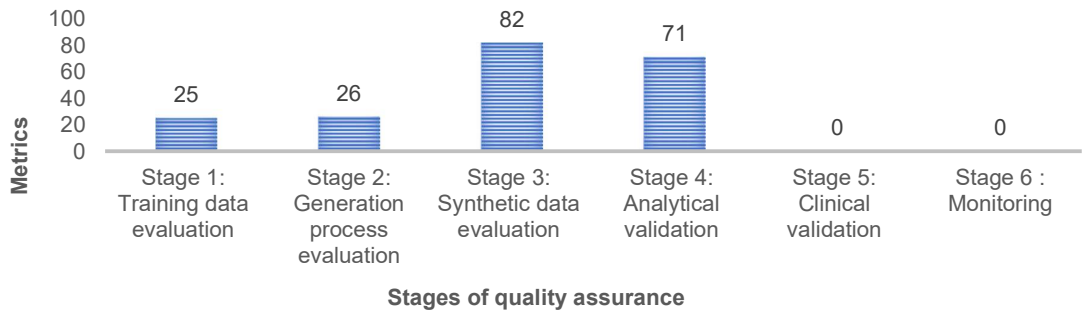

*Figure 6 The bar chart shows the proportion of the articles in the literature review that documented the use of quality assurance over the different stages of the process.*

24 of the 94 articles documented evaluating the training data (stage 1). All 24 documented pre-processing steps for record level data curation: normalization of variables, outlier detection and missingness. Four of the articles additionally documented handling class imbalance[50, 52-54]. In addition, one article documented de-identification of training data due to privacy risk[51], while two of the 94 articles[41, 50] evaluated case completeness – whether the original dataset represented an appropriate selection for the downstream use.

## 4  Discussion

### 4.1  Discussion of results

Our literature review reveals two emerging topics omitted from existing evaluation frameworks: "Fairness" and "Carbon Footprint".

Despite the growing emphasis on algorithmic fairness, the integration of fairness perspectives in articles was limited. As an emerging topic in the field, the 2023 NeurIPS SyntheticData4ML Workshop explicitly sought submissions on the topic of fairness (https://www.syntheticdata4ml.vanderschaar-lab.com/call-for-papers). Bhanot, Qi [31] demonstrated unfairness in SD generated from three commonly used public datasets: MIMIC, ATUS and ASGD Patients' Claims Dataset, underscoring the need for increased research in this area. We expect publications on fairness perspectives to continue increasing.

Carbon footprint measured through computational complexity was also rarely mentioned. While 9 out of the 94 articles used computational resources as an evaluation metric, none of the articles quantified the resource needs in terms of Carbon footprint.

For the responsible advancement of generative AI it is crucial to include environmental cost as a quality metric while encouraging further transparency and reporting [133]. Tools are available to assess emissions [134].

Traditional dimensions like representativeness and privacy were extensively covered in the articles of the review, and the results showed an overwhelming preference in applying statistical tests of distance or divergence (100 of the total 616 metrics).  For stress testing and scientific discoveries, SD generation should enrich datasets beyond being similar to the training data. Metrics choice must mirror the context for downstream use. Despite the importance of correlation between certain variables in the medical data, only a third of the similarity metrics focused on dataset patterns. For tabular data, the clinical logic in a temporal sequence may be important to capture, yet this aspect received less attention.

Interpreting metrics in the review posed challenges due to differing taxonomies and semantics. Although quality assurance was conducted by a second researcher, misunderstandings may have occurred. However, the overall trends in the data are considered clear and robust.

Around 90% of the reviewed articles focused on benchmarking generators rather than ensuring data quality for real-world applications. Consequently, most of these articles did not include assessments of original data completeness, and none addressed downstream clinical validation or monitoring. There may be a genuine lack of such studies unless this could be explained by too narrow search criteria or unreported data sources in clinical validation studies. The findings are consistent with the prevailing academic orientation of the field. To prepare for clinical applications of synthetic data, these additional stages should be integrated.

Benchmarking against the DNCR process demonstrates our framework aligns with real-world processes. It covers key stages and quality dimensions and provides transparency in metric choices and prioritization.

Structured according to concerns rather than how the metrics are calculated, our framework enhances comprehension for non-data scientist decision makers to scrutinize quality evaluations. Similarly, the high-level conceptual framework will keep its relevance as new metrics emerge and replace the old and can be adapted for other data modalities. The stages and dimensions can be applied across industries by replacing "clinical context" with "application context."

While the scope of our review was deep generative methods, the articles often compare several methods with little distinction in metric choices and our framework accommodates both deep and non-deep generative technologies.

## 4.2   Further work

The current absence of a framework for objective evaluation standards [14] leads to subjective considerations as a basis for sharing of SD. A consensus-backed framework within an organization would move decisions away from personal responsibility and provide assurance for safe and responsible sharing of SD to facilitate real-life application.

Standardizing terminology is essential, as demonstrated within our research team by varying interpretations of seemingly simple concepts like "framework" or "clinical effect". Mixed-competence teams promote comprehensibility and relevance to diverse users.

Our conceptual framework should maintain its relevance and grow alongside emerging generation methods and corresponding new evaluation techniques. Validation studies are needed to guide the choice of specific evaluation metrics in different contexts.

The demonstrated scarcity of fairness and carbon footprint metrics in the literature review calls for a collective effort. Our hope is that incorporation of these metrics in evaluations, the community can collectively steer toward a more ethical and sustainable advancement of AI.

# 5   Conclusion

Our synthetic data quality assurance framework promotes effective communication within and across disciplines by addressing the diverging terminology in existing frameworks. It goes beyond the conventional quality dimensions of similarity, usability, and privacy to address issues of bias amplification (fairness) and computational complexity and energy consumption (carbon footprint).

The framework promotes transparency in the priorities and trade-offs necessary to balance the quality aspects to assure the SD is fit for its intended purpose and is designed for real-life clinical application aligned with regulatory requirements. We demonstrate the applicability of our framework by using it to evaluate the DNCR breast cancer case. A unified approach for the evaluation of SD will facilitate the acceleration of safe and responsible implementation of digital innovation to aid healthcare delivery.

# 6 Summary table

**What is previously known on this topic:**

- Many evaluation metrics and frameworks for synthetic data have been suggested
- There are no agreed common framework or standards for quality assurance of synthetic data

**What this study added to our knowledge:**

- We demonstrate a gap in existing practice of quality assurance of synthetic data, missing important quality dimensions and steps to ensure safe real-life application
- We suggest a conceptual framework to enable safe real-life use of synthetic data, including relevant quality dimensions of similarity, usability, privacy, fairness and carbon footprint.

# 7 Authors' contributions

Author contributions according to the Contributor Roles Taxonomy:

Vibeke Binz Vallevik: Conceptualization; Data curation; Formal analysis; Funding acquisition; Investigation; Methodology; Project administration; Validation; Visualization; Writing - original draft; and Writing - review & editing.

Aleksandar Babic: Data curation, Formal analysis, Investigation; Methodology; Writing - original draft, Writing - Review & Editing

Serena E. Marshall: Data curation, Formal analysis, Investigation; Methodology; Writing - original draft, Writing - Review & Editing

Severin Elvatun: Data curation, Formal analysis, Investigation; Methodology; Writing - original draft, Writing - Review & Editing

Helga M. B. Brøgger: Data curation, Investigation, Writing - original draft, Writing - Review & Editing

Sharmini Alagaratnam: Investigation, Writing - Review & Editing, Supervision.

Bjørn Edwin: Investigation, Writing: Review & Editing, Supervision.

Narasimha Raghavan: Investigation, Data Curation, Writing-Review and Editing

Anne Kjersti Befring: Investigation, Writing - Review & Editing, Supervision.

Jan F. Nygård: Conceptualization, Investigation, Writing - Review & Editing, Supervision.

## 8 Acknowledgements

This study was partially funded by The Norwegian Research Council (grant number 333913). The funder played no role in study design, data collection, analysis and interpretation of data, or the writing of this manuscript.

We want to thank the Daan Knoors and the team at Netherlands Comprehensive Cancer Organisation (IKNL) for their interest in and their time contributing to our work with the DNCR case.## 9 Data Availability

The SD from the Dutch breast-cancer cohort is available at their website of the IKNL (https://iknl.nl/en/ncr/synthetic-dataset). The findings and data supporting this literature review are derived from the authors' insights and references cited within the paper.

## 10 Declaration of Competing Interest

All authors declare no financial or non-financial competing interests.

## 11 Declaration of Generative AI and AI-assisted technologies in the writing process

During the preparation of this work the authors used Chat GPT 3.5 on certain parts of the text for suggestions on how to shorten the text to improve readability. After using this tool, the authors reviewed and edited the content as needed and takes full responsibility for the content of the publication.

PAGE 26

# Appendices

# Appendix A. Benchmarking the framework with a case from the Dutch National Cancer Registry

This appendix summarizes the considerations made in the Dutch breast-cancer case and the reasoning behind choices made.

In the Netherlands, researchers can apply for access to data from the DNCR by submitting a data request to the Netherlands Comprehensive Cancer Organisation (IKNL). Completing such a request requires a certain level of medical expertise and knowledge about the data items available to be able to determine which are needed to answer the proposed research question. To stimulate a broader use, IKNL released a synthetically generated dataset based on a breast cancer cohort.

**Stage 0: Priorities and context**

The intended use was not a specific use case, rather to share a generic dataset showcasing what kind of data is available, but also to allow for some initial exploration and software development without the need to see real patient records. Making this knowledge more easily available will in turn stimulate more targeted use of the (real) data.

Previously, the typical user of the cancer registry data has been limited to specialists who have prior knowledge of the material. The SD is made openly available for other groups in the hope is that it will inspire researchers and data scientists from other domains to find new use for the data. Software that has been built using the SD should be possible to use directly on the real data, without any developer needing to access the real data with the increased privacy risk this gives.

As a national health registry, privacy was seen as a non-negotiable criterion for the generation and sharing of SD. The data should be structurally similar to the real data and preserve some of the statistical patterns, but without the risk of disclosing sensitive information about the original data subjects. Moreover, the SD was not designed for a specific use case and the webpage for requesting the SD states that it cannot be used for any form of clinical decision-making or policymaking, as analysis results might deviate from the actual data. Hence, some statistical similarity and usability requirements were relaxed in favour of providing strong privacy guarantees.

**Stage 1: Original data quality evaluation**

A team of breast cancer experts were consulted about the relevant data columns to be included in the dataset. The department at DNCR that processes data requests routinely do data quality assessments. The original dataset was pre-processed and de-identified prior to generating the SD to reduce privacy risk. For example, rare cases, patients below 18 years, minority groups and other outliers were removed. Looking at the frequency of each column, some continuous parameters were grouped into categories (less granularity, lower privacy risk) f. ex. age groups. In addition, the table was reduced to one incident (tumour) per patient to limit reidentification risk.

Bias was not considered. Low minority groups were removed rather than unsampled as a privacy risk reduction. Missing values were not removed, since these are statistical patterns that they seek to preserve to make the SD as realistic as possible.

**Stage 2: Generation performance evaluation**

In the choice of an appropriate generation method, the following was taken into account:

Data modality (tabular, not sequential), Listed limitations on data dimensionalities in terms of the maximum number of records and columns, and whether it was optimised for a small or large datasets. A not too heavy computational complexity was favoured, both in terms of manual labour and pre-processing needs, but also in computational power to be able to run multiple experiments with different settings and compare the results. Bias was not considered. The team tested various hyperparameter settings but mostly used recommended settings for hyperparameters (from the literature).

To choose the right generator, the team first looked at published articles to identify promising generators, discussed these with experts and ultimately experimented with selected models, evaluating similarity (Jensen-Shannon) and privacy levels (epsilon for DP).

Privacy was one of the strongest criterions for decision. The team wanted to use differential privacy (DP) to guarantee privacy. With the existing methods at the time it was a challenge to find methods that could also provide the required similarity and usability under strict privacy guarantees. The team experimented with histogram representations, Bayesian networks and GANs. The choice ultimately fell on Bayesian networks due to their ability to provide the required utility with strong privacy guarantees. A set of epsilon values were tested within a recommended range according to other privacy literature. Finally, a value for epsilon was chosen that was able to provide the required usability on various usability metrics.

**Stage 3: SD evaluation**

The statistical similarity of the generated data was evaluated comparing variables (univariate) and multivariate (comparing 2 or more variables.

Univariate comparison was based on visual distribution plots and a calculation of the Jensen-Shannon distance and average-variation distance.

For categorical variables (f. ex. Age groups), frequencies were compared. For continuous variables, density plots were used (f. ex. size of tumour, follow up time, etc).

Multivariate comparison was done with pairwise correlation plots. Joint distributions between columns were evaluated with visual plots and JS divergence.

As the data was not meant to be used for clinical decision-making, the team did not aim to make synthetic patients realistic. Also, the requirement for comprehensiveness played a part, where they wanted to include virtually all variables that were commonly requested to fully display what was available (resulting in around 45 variables). Hence, records become so unique that it was not desired to fully replicate them in the synthetic data.

As differentially private algorithms were used - which provide mathematical guarantees of privacy - the team did only a small number of post-generation privacy tests, for example by making sure that the distance between synthetic and original patients was large enough.

Bias was not considered in the SD.

Feedback from students using the dataset and comparing to a dataset from USA has been that the SD was lacking in clinical realism. A student reached out as he did an analysis of the SD and compared results with statistics, he found in academic literature that were done on US breast cancer data. Here he concluded that the results from the same analysis performed on the SD deviated quite a bit from what he saw in the US paper.

### Stage 4: Analytical validation

The downstream benchmark tests were not so decisive. A handful of generators from the academic literature that seemed to work well for similar datasets were implemented and tested using various real-world datasets and continued with the ones that worked best on datasets similar to the NCR.

Downstream performance was measured by looking at analyses techniques that are typically used with cancer data, e.g. ML prediction tasks, survival analysis etc and see how the results compare between the synthetic and the original. The team trained different ML models on both original and SD, then tested on a hold-out dataset from the real data that was not used to train the SD generator.

To evaluate results of prediction tasks, they used efficacy plots, ROC curves and calculated AUC. The results from these were very dependent on what variable they tested for. **On the target variable, where some variables managed to retain their statistical relationships well, others were less well-preserved.**

The team did log rank tests and survival analysis with Kaplan-Meier plots comparing the two datasets showed more variance in the SD. They made different KM curves by splitting survival time for each variable and see how well the curves aligned the original.

**Stage 5 and 6: Clinical validation and Monitoring**

As there was no specific downstream use defined, clinical validation of the trained AI tool and monitoring of the implemented AI tool was not relevant in this case.

# Appendix B. Additional results from the literature review

*Table B. 1 Summarizing the metrics count for different subgroups.*

| Quality metrics | Metrics (Articles) |
|---|---|
| **Similarity** | |
| Univariate level similarity: | |
|     Distribution coverage | 153 |
| Multivariate level similarity: | |
|     Distribution coverage | 47 |
|     Dataset structure | 88 |
| Sum similarity metrics: | 288 (78) |
| **Usability** | |
| Clinical logic | 11 |
| Generator performance | 17 |
| Downstream performance Benchmark | 190 |
| Sum usability metrics: | 218 (76) |
| **Privacy** | |
| Privacy risk (training data) | 1 |
| Privacy budget | 7 |
| Privacy loss | 27 |
| Privacy risk | 33 |
| Sum privacy metrics: | 68 (42) |
| **Fairness** | |
| Training data bias detection | 5 |
| Generator fairness evaluation | 0 |
| Fairness evaluation of SD | 2 |
| Downstream fairness evaluation | 3 |
| Sum fairness metrics: | 10 (6) |
| **Carbon footprint** | |
| Generic preprocessing | 22 |
| Generation prep processing | 1 |
| Computational cost | 9 |
| Sum carbon footprint metrics: | 32 (28) |

*Table B. 2*

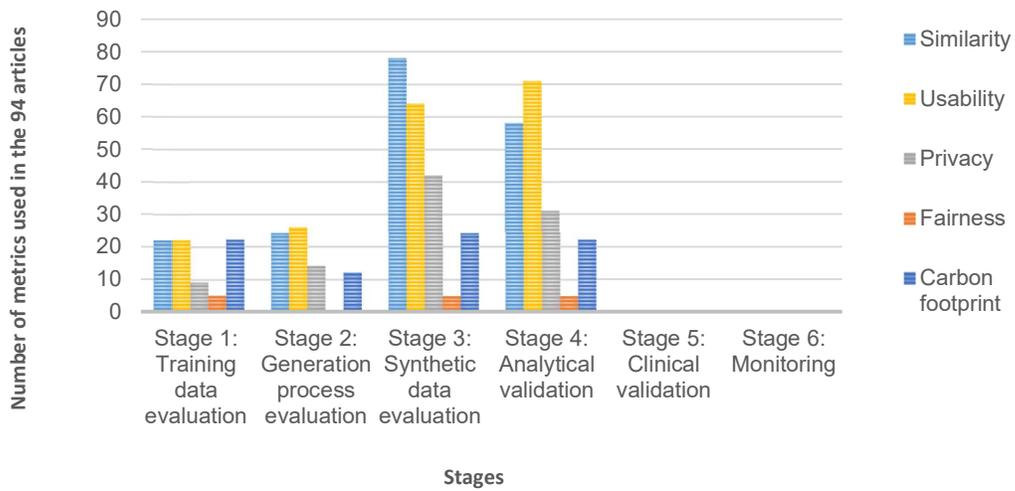

*Figure B. 1 Number of quality metrics used pr stage in our five quality dimensions registered in the review.*

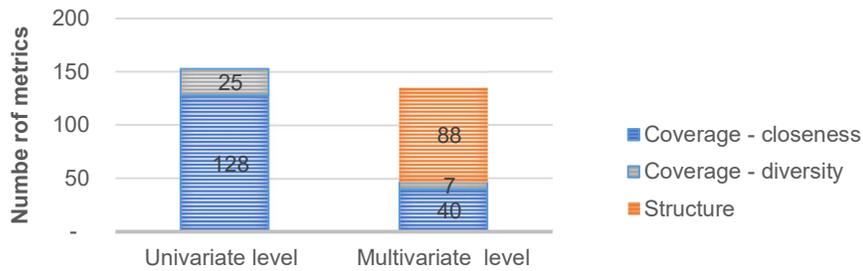

*Figure B. 2 Break-down of the type of subgroups of similarity metrics registered in the review.*

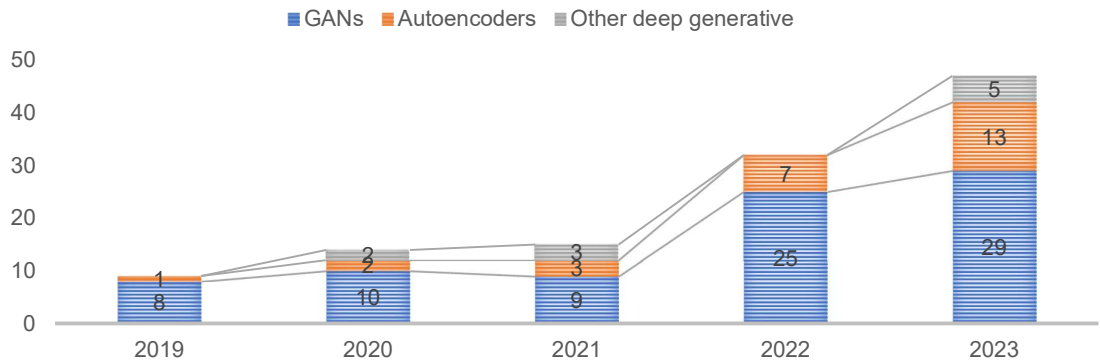

*Figure B. 3 Types of deep generative methods found in the literature review. (cut-off mid-October 2023)*

## Appendix C. Glossary

Glossary of specific terms used in this article are included here.

**Analytical validation:** Testing performance of a ML model with real data in an analytical setting.

**Case completeness**: The population in the dataset must be representative and relevant, preferably diverse enough to yield robust results.

**Clinical validation:** Testing the performance of a ML model with real data in a clinical setting to demonstrate that the device will work as intended, for the target population and in the clinical context.

**Deep generative generation methods:** Methods for generating synthetic data that apply deep learning models.

**Distance versus divergence metrics:** A *(distance) metric* or *distance function* is a function *d(x,y)* that defines the distance between elements of a set. It provides a way to measure how close two elements (such as numbers, vectors, matrices or arbitrary objects) are. Such function is required to satisfy the following conditions: (1) $d(x,y) \geq 0$ (non-negativity), (2) $d(x,y) = 0$ if and only if x=y, (3) $d(x,y) = d(y,x)$ (symmetry), and (4) $d(x,z) \leq d(x,y) + d(y,z)$ (triangle inequality). A *statistical distance* quantifies the distance between two statistical objects such as two random variables, or two probability distributions or samples, or the distance can be between an individual sample point and a population or a wider sample of points. A distance between populations can be interpreted as measuring the distance between two probability distributions. Statistical distance measurements do not usually abide by the principles of a classical metric (e.g., they may not be symmetrical). Certain kinds of distance measurements, which are an extension of squared distance, are known as (statistical) *divergences*.

**Downstream:** The data application part of the generation process

| | |
|---|---|
| **Generative models:** | Synthetic samples generated from real data are obtained by creating a model that captures the properties (distribution, correlation between variables, etc.) of the real data. Once the model is created, it is used to sample synthetic data. |
| **Upstream:** | The data production part of the generation process |
| **Record completeness:** | Whether the dataset is technically suitable for use. |
| **Variable:** | A variable refers to an individual measurable property or characteristic of a phenomenon. Variables can be continuous (numerical values that can take on any value within a certain range), or categorical (data that can be divided into distinct categories, such as colors). In tabular data, a column (e.g., age or gender) typically represents one variable. Variables are also referred to as features or attributes. |